\documentclass[letterpaper]{article} 
\usepackage{aaai25}  
\usepackage{times}  
\usepackage{helvet}  
\usepackage{courier}  
\usepackage[hyphens]{url}  
\usepackage{graphicx} 
\usepackage{natbib}  
\usepackage{caption} 
\usepackage{algorithm}
\usepackage{algorithmic}
\usepackage{mathrsfs}
\usepackage{nccmath}
\usepackage{subcaption}
\usepackage{booktabs}
\usepackage{xcolor}
\usepackage{graphicx}
\usepackage{amsmath}
\usepackage{amssymb}
\usepackage{nicefrac,xfrac}
\usepackage{pifont}
\usepackage{etoolbox}
\usepackage{tabulary}
\usepackage{mathrsfs}
\usepackage{nccmath}
\usepackage{needspace}
\usepackage{arydshln}
\usepackage{tabularx,colortbl}
\usepackage{tikz}
\usepackage{multirow, makecell}

\usepackage{newfloat}
\usepackage{listings}
\DeclareCaptionStyle{ruled}{labelfont=normalfont,labelsep=colon,strut=off} 
\floatstyle{ruled}
\newfloat{listing}{tb}{lst}{}
\floatname{listing}{Listing}
\newcommand{\ie}{
    \textit{i.e.}
}
\newcommand{\eg}{
    \textit{e.g.}
}

\title{FoMo: Multi-Modal, Multi-Scale and Multi-Task Remote Sensing Foundation Models for Forest Monitoring}

\author {
    Nikolaos Ioannis Bountos\textsuperscript{\rm 1,2,3},
    Arthur Ouaknine\textsuperscript{\rm 1,4},
    Ioannis Papoutsis\textsuperscript{\rm 2},
    David Rolnick\textsuperscript{\rm 1,4}
}
\affiliations {
    \textsuperscript{\rm 1} Mila - Qu\'ebec AI Institute\\
    \textsuperscript{\rm 2}Orion Lab, National Observatory of Athens \& National Technical University of Athens\\
    \textsuperscript{\rm 3} Harokopio University of Athens\\
    \textsuperscript{\rm 4} McGill University

}

\begin{document}

\maketitle

\begin{abstract}
Forests are vital to ecosystems, supporting biodiversity and essential services, but are rapidly changing due to land use and climate change.
Understanding and mitigating negative effects requires parsing data on forests at global scale from a broad array of sensory modalities, and using them in diverse forest monitoring applications. 
Such diversity in data and applications can be effectively addressed through the development of a large, pre-trained foundation model that serves as a versatile base for various downstream tasks.
However, remote sensing modalities, which are an excellent fit for several forest management tasks, are particularly challenging considering the variation in environmental conditions, object scales, image acquisition modes, spatio-temporal resolutions, \textit{etc}.
%
%
With that in mind, we present the first unified \textbf{Fo}rest \textbf{Mo}nitoring \textbf{Bench}mark (FoMo-Bench), carefully constructed to evaluate foundation models with such flexibility. FoMo-Bench consists of 15 diverse datasets encompassing satellite, aerial, and inventory data, covering a variety of geographical regions, and including multispectral, red-green-blue, synthetic aperture radar and LiDAR data with various temporal, spatial and spectral resolutions. FoMo-Bench includes multiple types of forest-monitoring tasks, spanning classification, segmentation, and object detection. 
To enhance task and geographic diversity in FoMo-Bench, we introduce \textbf{TalloS}, a global dataset combining satellite imagery with ground-based annotations for tree species classification across 1,000+ categories and hierarchical taxonomic levels.
Finally, we propose \textbf{FoMo-Net}, a pre-training framework to develop foundation models with the capacity to process any combination of commonly used modalities and spectral bands in remote sensing.
This work aims to inspire research collaborations between machine learning and forest biology researchers in exploring scalable multi-modal and multi-task models for forest monitoring and beyond. 
\end{abstract}

%
\begin{links}
    \link{Code and Data}{https://github.com/RolnickLab/FoMo-Bench}
\end{links}
\section{Introduction}
\label{sec:intro}

Forests are an essential part of Earth's ecosystems and natural systems, providing critical services on which humanity depends. They also play a vital role in biodiversity conservation, harboring countless species and maintaining ecological balance. However, forests are rapidly changing as a result of land use decisions, climate change and invasive species \cite{bonan_forests_2008, curtis_classifying_2018, hartmann_climate_2022} 
 \begin{figure*}

    \centering
    \includegraphics[width=0.96\linewidth]{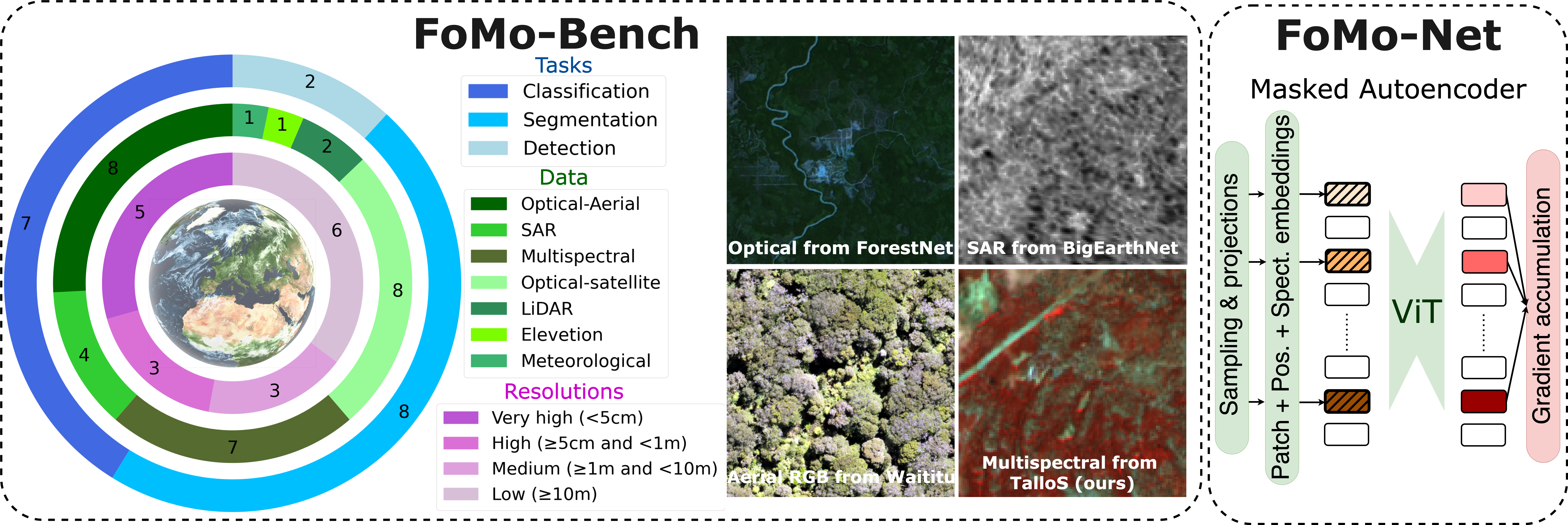}
    \captionof{figure}{FoMo-Bench evaluation framework and the FoMo-Net pretraining framework for foundation models.}
    \label{fig:fomo_teaser}
\end{figure*}
leading to significant biodiversity decline and economic losses. 
However, effective forest management and conservation are hampered by a lack of information, highlighting the need for scalable forest monitoring.
This gap encompasses multiple specific problems, such as tree species identification, tree size assessment, land cover classification, \textit{etc.}, which also vary considerably across geographies, climatic zones and landscapes. 

Remote sensing (RS) has been explored for forest monitoring using various sensors and spatial scales \cite{guimaraes_forestry_2020, michalowska_review_2021}, while increasingly, deep learning methods have been used to address various specific forest-related problems \cite{kattenborn_review_2021}. 
%
Simultaneously, the field of computer vision has seen a growth in foundation models \cite{bommasani_opportunities_2021}, \textit{i.e.} models flexible to diverse modalities, scales and data regimes while performing well across downstream tasks \cite{kirillov2023segment, jaegle_perceiver_2021, jaegle_perceiver_2022}.
However, it has been noted that general-purpose foundation models in computer vision may be ineffective in specific problem domains \cite{rolf2024mission}, presenting a need for foundation models explicitly designed for specific problem spaces, such as Earth observation \cite{lacoste2021toward} and weather or climate modeling \cite{mukkavilli2023ai}.
%
Foundation models offer the potential for a flexible backbone that can rapidly be fine-tuned on smaller annotated datasets. Given the diverse nature of RS data and applications, Earth observation foundation models should be able to operate on a wide array of Ground Sampling Distances (GSD), spectral bands, image acquisition modes, and environmental conditions. Assessing the capacity of such models is far more challenging than comparing performance on standard RS datasets. 
In this work, we propose a framework to evaluate general remote sensing foundation models utilizing the inherent diversity of critical forest-monitoring tasks.
%
Our \textbf{Fo}rest \textbf{Mo}nitoring \textbf{Bench}mark (\textbf{FoMo-Bench}) brings together a wide range of publicly available datasets for forest monitoring, encompassing significant spatial, temporal, and spectral diversity as illustrated in Fig.~\ref{fig:fomo_teaser}. 
FoMo-Bench includes a mixture of classification, segmentation and detection tasks, from land cover classification to tree crown segmentation.
To improve the geographic diversity of FoMo-Bench, we also create \textbf{TalloS}, 
global, multi-modal dataset focused on tree species identification based on worldwide forest inventories \cite{jucker2022tallo}.
Each sample included in TalloS combines data acquired from the Sentinel-2 and Sentinel-1 satellite missions, meteorological data from the ERA-5 \cite{hersbach_era5_2020} dataset, elevation information, and georeferenced tree species information 
for multi-label classification.

Finally, we develop \textbf{FoMo-Net}, a pretraining framework which can flexibly process most commonly used modalities and spectral bands in RS with a single backbone. \textbf{FoMo-Net serves as an initial response to an ambitious research question}: Can we unify foundation models trained on specific individual remote sensors and GSD into a singular, versatile foundation model applicable and adaptable to the requirements (\textit{e.g.} combination of spectral bands, GSD, and geographic location) of a wide range of downstream tasks? To put FoMo-Net to the test, we utilize FoMo-Bench combined with unlabeled data archives to develop an initial multi-modal RS foundation model exploiting the FoMo-Net algorithm with a simple vision transformer (ViT)~\cite{dosovitskiy2020image} encoder.
 Following common practice in RS foundation models, we assess the capacity of our foundation model by finetuning to the FoMo-Bench tasks. 
  While supervised baselines in FoMo-Bench set a high performance bar, FoMo-Net delivers strong predictive accuracy handling a wide array of input sensors, spectral band combinations, GSD and forest monitoring tasks. It does not require any change in the encoder, marking a crucial first step toward a multi-task, variable-modality model for forest monitoring, stimulating progress in this nascent field and laying the groundwork for future foundation models. 
 We elaborate more on our findings and potential lines of research in Sec.~\ref{sec:discussion}.
Overall, our contributions can be summarized as follows: 
\begin{itemize}
    \item We provide FoMo-Bench the first, curated benchmark for forest monitoring, with a global spatial coverage and a diverse set of evaluation tasks;
    \item We introduce TalloS, a novel multi-modal dataset for tree species multi-label classification;
    \item We introduce FoMo-Net, a unique pre-training approach designed to train foundation models with great flexibilty in terms of the input bands, their ground sampling distance, and geographic location. 
\end{itemize}

\section{Related Work}
\label{sec:related_work}

Large-scale RS data has been available for many years through publicly funded projects such as the Sentinel and Landsat missions.
Building on such sources, various machine learning benchmarks have been developed, such as BigEarthNet-MM \cite{sumbul2021bigearthnet}, Sen12MS \cite{schmitt2019sen12ms},  Sen4AgriNet \cite{sykas2022sentinel}, and CropHarvest \cite{tseng2021cropharvest}. 
These initiatives have facilitated extensive research, particularly in land use and land cover (LULC) management and forest-related tasks such as tree species identification \cite{fassnacht_review_2016}. 

\textbf{Forest monitoring:} High resolution image recordings from unmanned aerial vehicles (UAVs) are part of recent datasets for automated canopy mapping \cite{galuszynski2022automated}, woody invasive species detection \cite{kattenborn2019uav} and tree species detection \cite{kattenborn2020convolutional}. In particular, ReforesTree \cite{reiersen2022reforestree} combines UAV recordings with inventories for both detecting trees and estimating their aboveground biomass.
The NeonTree dataset \cite{weinstein2020benchmark} is unique in grouping airborne recordings including light detection and ranging (LiDAR), red-blue-green (RGB) and hyperspectral measurements.
Although inventories have been used alongside satellite data in Russia with less than a thousand samples \cite{brieger_advances_2019}, we are not aware of any RS dataset providing fine annotations on a large volumetric and geographical scale. 

\textbf{RS benchmarks:} Benchmarks across the space of computer vision have moved beyond single tasks to encompass a diversity of related tasks. The large-scale BigEarthNet-MM dataset, built from Sentinel-1 and Sentinel-2 data, is designed to assess both performance and training complexity of deep learning methods \cite{papoutsis2023benchmarking}.
The SustainBench \cite{yeh2021sustainbench} combines various satellite imagery and street-level images to monitor sustainable development goals covering 15 different topics.
Geo-Bench \cite{lacoste2023geo} proposes to group RS datasets -- including SAR, multispectral, elevation and cloud probabilities data -- to tackle six classification and six segmentation tasks focused on LULC. However, Geo-Bench's limited spatial coverage reduces its potential for broad global applicability.
There remains no benchmark focused on forest monitoring worldwide, combining the full suite of forest inventories, aerial and satellite data, and including the several relevant classification, segmentation and detection tasks.

\textbf{RS foundation models:} Foundation models have been investigated for RS datasets
\cite{sun_ringmo_2022, cong2022satmae}, grounded mainly on LULC related tasks.
While these approaches have shown initial success, they are often limited to specific satellite products and lack diversity in spectral bands and GSD. 
To address the GSD limitation, Scale-MAE \cite{reed2023scale} was trained on the FMoW-RGB dataset \cite{christie2018functional}, adopting a masked image modeling (MIM) approach. This model incorporated GSD positional encoding and a Laplacian-pyramid decoder to enforce multi-scale feature learning.
Similarly, Cross-Scale MAE \cite{tang2023crossscale} leverages FMoW-RGB to learn scale-invariant representations by integrating contrastive loss in the MIM task and introducing a cross-prediction task, where lower-resolution images predict the higher-resolution ones.
SpectralGPT \cite{hong_spectralgpt_2023}, utilizes the spectral diversity of Sentinel-2 by employing 3D tokens, which presents scalability challenges when dealing with a large number of bands. 
SkySense \cite{guo2023skysense} stands out as the largest RS foundation model to date with 2 billion parameters. It relies on three aligned modalities \ie WorldView RGB, Sentinel-2 multispectral and Sentinel-1 radar data, though this limits its applicability to a broader range of RS sensors. SkySense pretraining utilizes contrastive learning and cross-modal alignment. In a similar fashion, OmniSAT \cite{astruc2024omnisat} and CROMA \cite{fuller2023croma} also explore contrastive and reconstruction losses to efficiently fuse information from spatially aligned RS imagery. 
Both models use Sentinel-1 and Sentinel-2 data, with CROMA focusing on single timestamps and OmniSAT incorporating time series and high resolution aerial data.
USatMAE \cite{ray2016usat} employs paired NAIP and Sentinel-2 data, using dedicated patch embeddings for each spectral band and a spectral group pooling strategy that groups bands by spatial resolution. These tokens are then processed by a transformer trained via masked image modeling. 
Similarly, SpectralMAE \cite{zhu2023spectralmae} focuses on paired hyperspectral data, using a transformer encoder with masking and attention mechanisms applied exclusively to the spectral dimension.
The recently introduced Presto algorithm \cite{tseng2023lightweight} proposes a MIM framework for paired data, with a built-in functionality to process a variable number of spectral bands, but is restricted to single-pixel time series. 
Finally, DOFA \cite{xiong2024neural}, aims to create a sensor agnostic transformer encoder through MIM, utilizing a wavelength-conditioned dynamic patch embedding which provides flexibility regarding input spectral bands.

Recently, there has been a surge of interest in flexible yet general RS foundation models that are capable of handling diverse data sources and solving a wide range of downstream tasks.
In this work, we aim to contribute to this line of research by: a) introducing an evaluation framework designed to assess the generalization ability of future models across various Earth observation modalities and conditions, with a specific focus on forest monitoring tasks that have high ecological, environmental, societal and economic impact, and b) proposing a novel pretraining paradigm that leverages any number of unaligned modalities to create sensor-agnostic foundation models. To the best of our knowledge, foundation models have yet to be trained on such a diverse range of inputs or applied to multiple downstream tasks at varying resolutions (from a few centimeters to tens of meters), while incorporating a wide array of spectral bands. This approach addresses a significant gap in the field, as highlighted by recent studies \cite{ouaknine2023openforest}.

\section{FoMo-Bench}
\label{sec:fomo_bench}
\begin{figure*}[!t]
  \centering
  \begin{subfigure}{0.5\linewidth}
    \includegraphics[width=\linewidth]{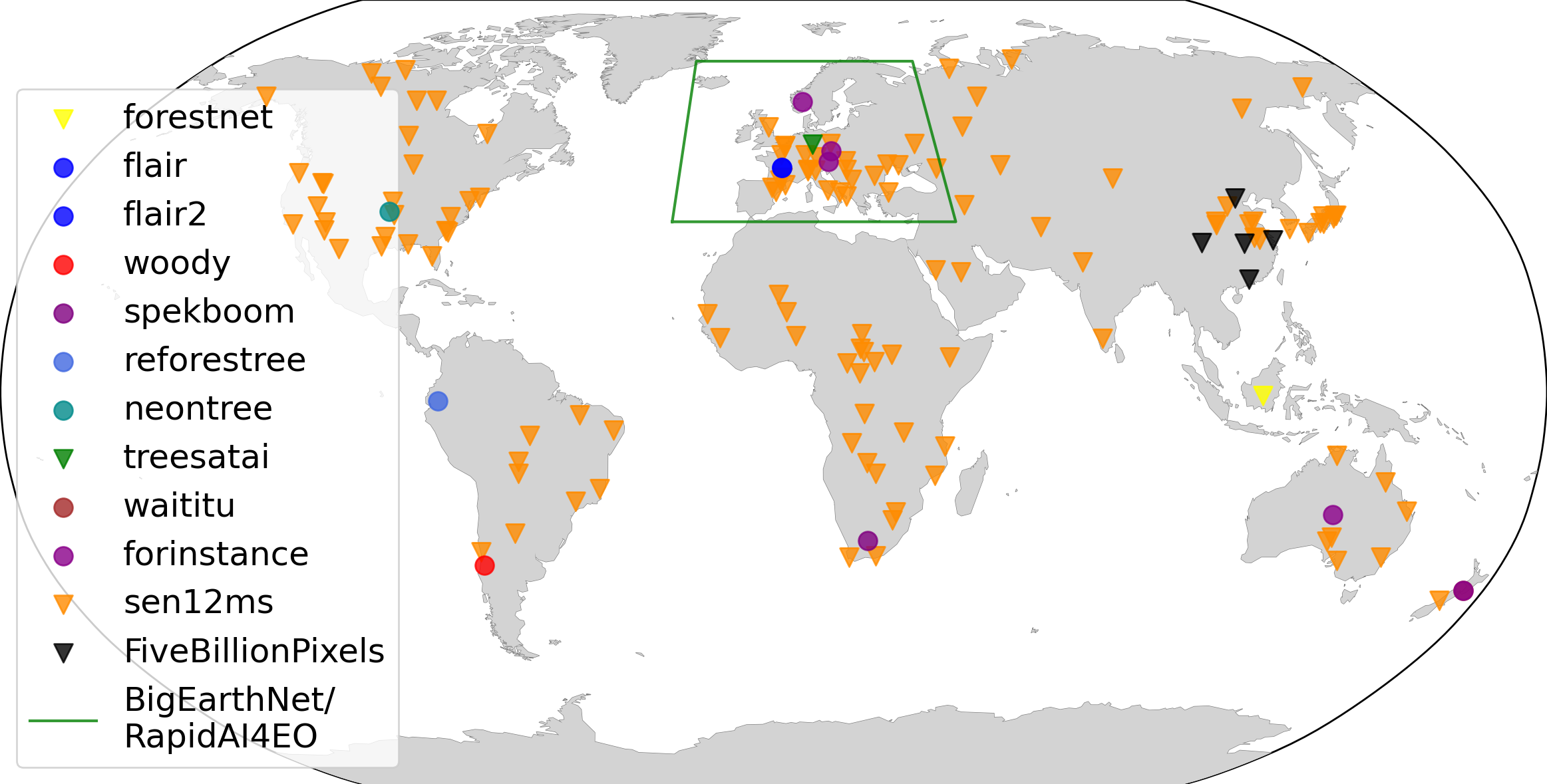}
    \caption{FoMo-Bench spatial distribution w/o TalloS.}
    \label{fig:fomo_distrib_wo_tallos}
  \end{subfigure}%
  \begin{subfigure}{0.5\linewidth}
  \centering
    \includegraphics[width=\linewidth]{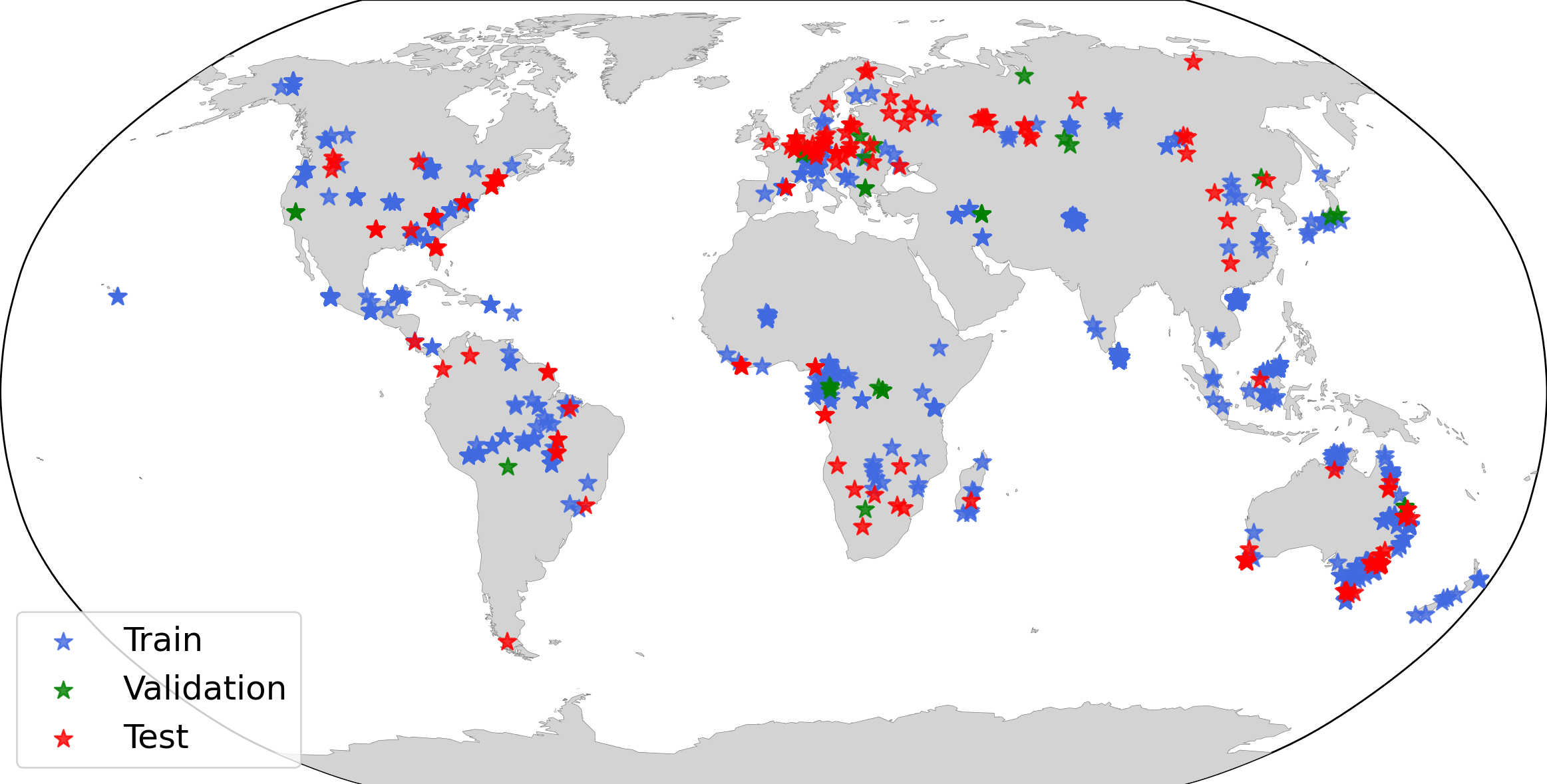}
    \caption{TalloS spatial distribution.}
    \label{fig:tallos_distr}
  \end{subfigure}
  \caption{\textbf{FoMo-Bench spatial distribution.} (a) Spatial distribution of the datasets included in FoMo-Bench, without TalloS; satellite-based datasets are marked with arrows, aerial-based datasets with circles. (b) Distribution of the train, validation and test splits of the TalloS dataset worldwide.
  }
  \label{fig:tallos_fomo_distribs}
\end{figure*}
We introduce FoMo-Bench, which offers a robust testbed for multimodal RS foundation models, grounded on the diversity of forest-monitoring tasks. Unlike the envisioned multimodal models, conventional methods in FoMo-Bench are task-specific, delivering strong application-optimized baselines though incapable of cross-task adaptation.
%
By carefully selecting datasets composing FoMo-Bench, we have ensured that it (1) encompasses a diversity of geographic locations and environmental conditions, (2) spans a range of task types, and (3) includes input data from a wide variety of sensors \textit{e.g.} optical, SAR, multispectral, DEM and LiDAR, as well as a range of ground sampling distances, from a few centimeters to $60\text{m}$ per pixel. 
The diversity of the FoMo-Bench data and tasks is illustrated in Fig.~\ref{fig:fomo_teaser} and described in more detail below.
We have curated all the datasets in FoMo-Bench, spatially split them to train machine learning models and created a reproducible environment for future work with a unified set of PyTorch data loaders. To our knowledge, this is the first benchmark grouping such diverse forest monitoring tasks both in terms of problems they address as well as in regards to the input data.

\textbf{Satellite-based datasets:} 
FoMo-Bench includes seven datasets involving satellite imagery, namely BigEarthNet-MM, Sen12MS, RapidAI4EO, ForestNet, FiveBillionPixels, TreeSatAI, and TalloS.
The \textbf{BigEarthNet-MM} and \textbf{Sen12MS} \cite{schmitt2019sen12ms} datasets are two of the most commonly used multi-modal datasets for LULC in RS. 
Both are based on the Copernicus Sentinel missions offering SAR and multispectral data from Sentinel-1 and Sentinel-2 respectively. For the BigEarthNet-MM dataset, we adapt the existing label structure, based on the Corine land cover database \cite{european_environment_agency_corine_2019}, to encompass 19 distinct land cover classes including broad-leaved, coniferous and mixed forests. While BigEarthNet-MM covers 10 countries in Europe, the Sen12MS dataset has a global spatial coverage (see Fig.~\ref{fig:fomo_distrib_wo_tallos}), offering annotations for 17 classes based on MODIS land cover maps \cite{friedl_modis_2010} with 5 different classes for forest cover. 
FoMo-Bench also includes \textbf{RapidAI4EO} \cite{rapidai4eo} as a third LULC dataset
with a unique source of high resolution satellite imagery offering timeseries of Planet data, at 3m resolution, for 500,000 locations across Europe. 
We adapt RapidAI4EO labels to encompass 24 classes, 12 of them relevant to forest monitoring. 
The \textbf{ForestNet} dataset \cite{irvin2020forestnet} aims to identify drivers leading to deforestation and uses Landsat-8 imagery.
\textbf{FiveBillionPixels} dataset \cite{FBP2023} provides high resolution Gaofen-2 images ($4\text{m}$) for LULC of cities in China with 24 categories including 5 forest monitoring classes.
%
The \textbf{TreeSatAI} dataset \cite{treesatai} offers multi-modal data (\textit{e.g.} aerial images, SAR and multispectral) for tree species identification, including 20 tree species derived from forest administration data in Lower Saxony, Germany.
Finally, we introduce a completely novel dataset, \textbf{TalloS}, detailed in Sec.~\ref{sec:tallos}. 


\textbf{Aerial-based datasets:} 
The limited spatial resolution offered by free satellite data is not suited for certain RS tasks, such as tree canopy detection or segmentation.
FoMo-Bench includes eight datasets with high resolution data from aerial sensor data, namely NeonTree, Woody, ReforesTree, Spekboom, Waititu, FLAIR \#1, FLAIR \#2 and FORinstance. 
The \textbf{NeonTree} dataset \cite{weinstein2020benchmark} focuses on tree crown detection with RGB imagery, LiDAR and hyperspectral data recorded from airborne systems. FoMo-Bench includes the NeonTree high resolution RGB imagery (0.1m resolution) along with point cloud LiDAR data. 
The \textbf{Woody} dataset \cite{kattenborn2019uav} is composed of RGB images recorded from an UAV, at several cm resolution, for detection of several invasive tree species in Chile. 
Similarly, the \textbf{Spekboom} \cite{galuszynski2022automated} and \textbf{Waititu} \cite{kattenborn2020convolutional} datasets are composed of RGB images recorded from UAVs to map tree species in South Africa and New Zealand, respectively. 
In the same vein, the \textbf{ReforesTree} dataset \cite{reiersen2022reforestree} is designed for tree crown detection with 6 classes using high-resolution RGB images recorded from an UAV, as well as ground measurements, such as tree height and diameter at breast height.
The \textbf{FORinstance} dataset \cite{puliti2023forinstance} provides a unique source of data with LiDAR point clouds and semantic annotations collected in 5 countries (see Fig.~\ref{fig:fomo_distrib_wo_tallos}).
Unlike the previous aerial datasets, \textbf{FLAIR~\#1} \cite{ign2022flair1, garioud2023flair} and \textbf{FLAIR~\#2} \cite{ign2023flair2, garioud2023flair} are large-scale datasets testing spatiotemporal generalization. 
Both datasets cover wide areas in France, providing high resolution aerial images (0.2m) and elevation maps (plus Sentinel-2 timeseries in the case of FLAIR~\#2). These datasets test LULC classification across 19 classes.

\section{The TalloS Dataset}
\label{sec:tallos}

As shown in Fig.~\ref{fig:tallos_fomo_distribs}, the spatial coverage of pre-existing datasets in FoMo-Bench is limited. We improve the diversity and spatial coverage of FoMo-Bench by designing a new dataset, TalloS. 
This dataset couples manual tree inventories with multispectral and SAR timeseries imagery for a fine grained multi-label classification challenge. 
 As described in the previous section, various forest monitoring tasks have relied on high-resolution UAV data, which, despite offering detailed insights, suffer from limited spatial coverage due to human and hardware constraints \cite{galuszynski2022automated, reiersen2022reforestree, kattenborn2019uav, kattenborn2020convolutional}. 
TalloS is a large-scale multi-modal dataset of satellite-based imagery for tree species multi-label classification.
It is based on the global Tallo database \cite{jucker2022tallo}, containing almost $500,000$ manual georeferenced recordings aggregating information from forest inventories including the tree species as well as tree height, stem diameter, and crown radius.
%
\begin{table}[!t]
\begin{center}\fontsize{7}{11}\selectfont
    \begin{tabular}{cccc}
    \toprule
      Dataset   & Sensors & Spatial res. & Sampling Weight \\ \toprule
       SSL4EO-Landsat & Landsat 8-9  & Low & 0.2   \\\hline
       RapidAI4EO  & Planet, S-2 & Medium, Low   &0.2  \\\hline
       TalloS  & S-1, S-2, DEM & Low &0.2\\\hline
       FLAIR \#1 &Aerial& High &  0.1 \\ \hline
       FiveBillionPixels & Gaofen-2 & Medium & 0.2\\ \hline
       UAV-datasets & Aerial & Very high &0.1\\ \hline
    \end{tabular}
    \end{center}
    \caption{\textbf{FoMo-Net pre-training datasets.} Our framework has been trained with four satellite-based datasets and four aerial-based datasets. 
    Their spatial resolution is defined as very high ($< 5\text{cm}$), high ($\ge 5\text{cm}$ and $< 1\text{m}$), medium ($\ge 1\text{m}$ and $< 10\text{m}$) or low ($\ge 10\text{m}$). The sampling weight is based on the frequency of each source in our pretraining datasets and is used for the modality sampling. S-1 and S-2 stand for Sentinel-1 and Sentinel-2 respectively.}
    \label{tab:pretraining_datasets}
\end{table}
Due to the modest GSD of publicly accessible satellites ($\approx 10\text{m}$), we design TalloS to concentrate on areas containing a minimum of 50 trees, as regions with lower tree populations lack distinctive signatures in satellite imagery. This choice balances diversity and strong signal for model prediction. There are $\approx335$k locations including 50 or more trees, but only $\approx251$k with 100 or more, reducing the dataset size by 25.1\% and the number of genera by more than 11\%. 
TalloS data represent a time series from 2016 to 2023 with a temporal frequency of 15 days. Each timestep consists of a multispectral image from Sentinel-2, a SAR image from Sentinel-1, a digital elevation model (DEM) from Copernicus GLO-30 at 10m per pixel resolution, along with 
 meteorological information from ERA-5.
 Sentinel-2 data is selected with cloud coverage below 30\%, and Sentinel-1 captures are aligned as closely as possible to Sentinel-2 acquisitions.
 Spectral bands with lower GSD are resampled to 10m per pixel for consistency with conventional models.
Each RS sample of TalloS is mapped to ground-validated and fine grained measurements of trees in its given area, including tree species, genus and family as well as their specific traits. For the purpose of this study, we focus on the genus level labels, as distinguishing between species may require higher resolution imagery. Throughout this paper, unless explicitly stated otherwise, we refer to the various genus categories of TalloS as its classes. 
TalloS represents a fine-grained classification problem involving $>$1,000 classes with a heavily skewed data distribution. The task is framed as multi-label classification, with the goal of predicting every tree genus that is depicted in the image timeseries.
In Fig.~\ref{fig:tallos_distr}, we illustrate the wide spatial distribution of TalloS as well as the proposed training, validation and test splits. 
Our goal is to maximize the geographic spatial coverage of each split while ensuring that (a) regions used for training and testing do not overlap, 
(b) most classes present in the validation and test sets are also included in the training set. Accommodating both conditions proved challenging due to the limited geographic distribution of many tree genera. We present TalloS samples along with its class distribution in Appendix A.

\section{FoMo-Net}
\label{sec:fomo-net}

Building on FoMo-Bench and inspired by the necessity for a generalized foundation model, we introduce FoMo-Net. 
FoMo-Net is a pretraining paradigm for learning to process all of the most common modalities in the RS domain with a single, sensor-agnostic foundation model. In this work we treat each spectral band of each sensor as a separate modality.
Our approach is driven by three core considerations. First, it should maximize flexibility to process various input settings, without relying on specific sets of sensory modalities.
Second, it should have the capacity to process information and generate meaningful representations for the whole globe.
%
Third, it should be applicable to a wide range of downstream tasks. 

\begin{figure}[t!]
   \centering
    \includegraphics[width=1.\linewidth]{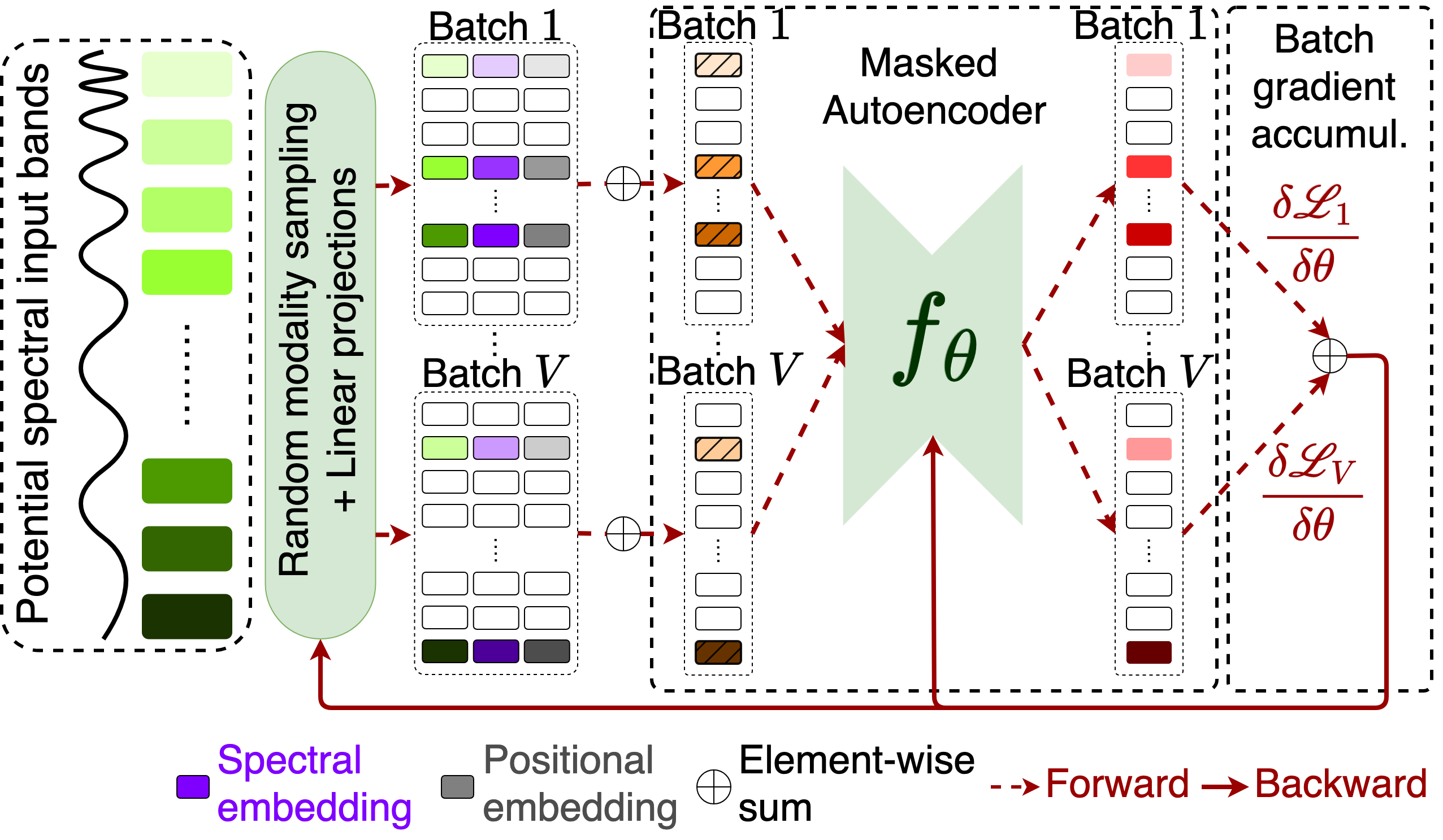}
    \caption{\textbf{FoMo-Net pre-training framework.} Considering a set of potential spectral input bands and a sub-training process of $V$ batches, each batch includes a sub-set of bands w.r.t. to the sampled dataset and modalities it contains. 
    For each batch, the input band embeddings, the spectral embeddings and the positional embeddings are summed element-wise to generate individual embeddings per band.
    The masked autoencoder, $f_{\theta}(\cdot)$ parameterized by $\theta$, reconstructs the partially masked inputs, illustrated with dashed squared, of a given batch. The gradients of the loss $\mathscr{L}(\cdot)$ are accumulated 
    and backpropagated through both $f_{\theta}(\cdot)$ and the linear projections. For the sake of clarity, we do not include the elevation modality which is not considered as a spectral band, but included in the pre-training scheme.}
    \label{fig:fomonet}
\end{figure}

\textbf{Pre-training data:}
FoMo-Net pre-training scheme contains rich multi-sensor information, both paired and unpaired, from most parts of the world. In particular, we use the RapidAI4EO dataset as a source of Planet and Sentinel-2 data; TalloS containing DEM, Sentinel-1, and Sentinel-2; SSL4EO-Landsat \cite{stewart2023ssl4eo} to acquire global information from Landsat 8 and 9; a combination of all available UAV datasets in FoMo-Bench; 
and the FiveBillionPixel dataset providing a unique source of Gaofen-2 high resolution satellite imagery. 
The datasets used in the FoMo-Net pre-training framework are detailed in Tab.  \ref{tab:pretraining_datasets}. 
%
The proposed framework is able to process any combination of the 36 most common modalities in RS, stemming from Sentinel-1, Sentinel-2, Landsat 8-9, Planet, Gaofen-2 and UAV sensors with ground sampling distance spanning from a few centimeters to $60\text{m}$ per pixel. 
%
In this work, a distinct modality is defined as any band with a unique combination of wavelength and spatial resolution.
Given its importance in Earth observation, digital elevation models (DEM) are also included in our definition as a specific band. 
The FoMo-Net pre-training pipeline is 
illustrated in Fig.~\ref{fig:fomonet} and will be detailed in the following sections. Additionally, the pseudocode of the pre-training pipeline is summarized in Appendix B. Point-cloud data are excluded from the pretraining set and initial FoMo-Net design due to the need for specialized processing, which is deferred to future work (Sec.~\ref{sec:discussion}).

\textbf{Approach to variable spectral bands:}
Let $\mathcal{D} = \{\mathcal{D}_1, \cdots, \mathcal{D}_{n}\}$ be a set of $n$ datasets, and $X = \{X_1, \cdots, X_{m}\}$ be a set of $m$ spectral bands, where $X_i \in \mathbb{R}^{H \times W}$ where $H$ and $W$ represent respectively the height and width dimensions.
At each iteration, a training batch $\mathcal{B}$ contains a variable number of spectral bands, each sampled with respective probabilities 
$\alpha_i \in \{\alpha_1, \cdots, \alpha_m\}$.
In our experiments, we set $\alpha_i$ proportional to the frequency with which the $i$-th band occurs in $\mathcal{D}$ (see Tab. \ref{tab:pretraining_datasets}).
Each band $X_i$ is tokenized into $N$ patches of size $P$ according to the transformation $s$, so that $s(X_i) \in \mathbb{R}^{N \times \lfloor \frac{H}{P} \rfloor \times \lfloor \frac{W}{P} \rfloor}$. The tokenized input $s(X_i)$ is then embedded using a linear transformation $t_i \in \{t_1, \cdots, t_m\}$ learnt during the optimization process as $t_i \colon \mathbb{R}^{N \times \lfloor \frac{H}{P} \rfloor \times \lfloor \frac{W}{P} \rfloor}  \rightarrow \mathbb{R}^{N \times d}$.
We considered two projection setups: The first (noted FoMo-Net$_{1}$) uses a single linear projection for all spectral bands, while the second (denoted FoMo-Net$_{m}$) projects each spectral band with its own linear projection. Our initial experiments suggest that FoMo-Net$_{1}$ yields significantly better results.
The input to our model, then, is provided as:
\begin{equation} \label{eq:embeddings}
    t_i(s(X_i)) \oplus \mathbf{S} \oplus \mathbf{P},
\end{equation} 
where $\mathbf{S} \in \mathbb{R}^{N \times d}$, $\mathbf{P} \in \mathbb{R}^{N \times d}$ are respectively the learnt spectral and positional embeddings, and $\oplus$ denotes the element-wise sum.

\textbf{Spectral band MAE:}
The next step of the procedure follows a masked autoencoding (MAE) framework \cite{he2022masked} that is shared by various high-performing foundation models for RS (see Sec.~\ref{sec:related_work}). 
 However, our method differs from a typical MAE in two key aspects (see Fig.~\ref{fig:fomonet}).

First, we introduce a random spectral band selection process by sampling in both $\mathcal{D}$ and $X$, representing the datasets and spectral bands, respectively. The number of selected spectral bands is randomly defined at each iteration. This approach promotes modeling interactions between a wide range of band combinations, creating a highly flexible encoder able to process any set of spectral bands.
%
Next, we incorporate both spectral and positional embeddings, allowing the model to capture information regarding the spatial position of the patch and its spectral band, as detailed in Eq.~\ref{eq:embeddings}. Both types of embeddings are trainable and randomly initialized.
%
Given the randomness in the input space, the spectral embedding is of utmost importance as no other information regarding the nature of the input is provided. Finally, as each band is tokenized independently the transformer is able to apply the attention mechanism to both spectral and spatial dimensions.

\textbf{Batch gradient accumulation:}
The combination of the proposed modality sampling and token masking is particularly computationally efficient in contrast to the exhaustive generation and processing of tokens from the entire spectrum of information. However, randomly choosing datasets and spectral bands at each training iteration results in highly heterogeneous batches, potentially destabilizing the optimization process.
To address this, the FoMo-Net pre-training algorithm exploits gradient accumulation (see Fig.~\ref{fig:fomonet}). 
Letting $f_\theta$ be a neural network parameterized by $\theta$, and $\mathscr{L}$ be a loss function, the gradient is accumulated during $V$ forward passes according to $\sum_{v=1}^{V} \nicefrac{\delta \mathscr{L}_v (f_{\theta}(\mathcal{B}^{v}))}{\delta \theta}$, where $\mathcal{B}^{v}$ is the $v$-th batch of input data (see Eq.~\ref{eq:embeddings}), and it is this sum that is backpropagated through $f_\theta$.

%
\textbf{FoMo-Net backbone:}
In this work we apply FoMo-Net on a ViT encoder with 12 layers, and 12 attention heads. 
We limit the maximum number of sampled spectral bands at the pre-training stage to 4 and the pixel resolution to $64\times64$ for efficiency and train for 300 epochs. 

\textbf{Modelling challenges:}
FoMo-Net must account for interactions across a wide range of spectral bands potentially useful for downstream tasks, unlike standard foundation models tailored to specific sensors. Additionally, varying GSD requires capturing features and patterns at multiple scales, some of which are a) underrepresented and b) lack sufficient data points with wide spatial coverage. Finally, given the diversity of information in the input and the large-scale pretraining dataset, scaling up the backbone could further improve performance. However, this would require significantly more computational resources.

\begin{table*}[!t]
\begin{center}\fontsize{8}{11}\selectfont
%
\begin{tabular}{c c c l l c c }
\toprule
Dataset & Modalities & \#C & Method & \#P (Mil.) &F1 micro & F1 macro \\
\midrule
\multirow{6}{*}{\multirowcell{2}{\textbf{BigEarthNet} \\ \cite{sumbul2021bigearthnet}}}
& \multirowcell{6}{Multispec.\\ SAR}
& \multirow{6}{*}{\textbf{19}}
& ResNet50 & 23.58 & 77.18 \scriptsize ($\pm0.32$) & 70.32 \scriptsize ($\pm1.32$) \\
& & & ViT & 87.97 & \textbf{78.09} \scriptsize ($\pm0.34$) &  \textbf{72.29} \scriptsize ($\pm0.61$)\\
& & & ConvNext & 87.60 & \underline{77.25} \scriptsize ($\pm0.26$) & \underline{69.66 }\scriptsize ($\pm0.73$)\\
& & & PolyNet$_{\text{in}}$ & 0.038 &73.51 \scriptsize ($\pm0.42$) & 66.07 \scriptsize ($\pm0.20$)\\ 
\cdashline{4-7}
 & & & FoMo-Net$_{1}$ & 66.39 & 73.42 \scriptsize($\pm0.10$) & 61.36 \scriptsize($\pm0.55$)\\
& & &FoMo-Net$_{m}$ & 73.57& 66.11 \scriptsize($\pm0.07$) & 50.0 \scriptsize ($\pm0.25$)\\
 \hline
\multirow{6}{*}{\multirowcell{2}{\textbf{Sen12MS} \\ \cite{schmitt2019sen12ms}}} 
&  \multirowcell{6}{Multispec.\\SAR}
& \multirow{6}{*}{\textbf{17}}
& ResNet50 & 23.57 &51.34 \scriptsize ($\pm1.71$) & \underline{32.62} \scriptsize ($\pm0.33$) \\
& & & ViT &87.97 & \underline{51.49} \scriptsize ($\pm1.24$) & 31.77 \scriptsize ($\pm1.72$)\\
& & & ConvNext &87.60 &50.86 \scriptsize ($\pm3.91$) &30.02 \scriptsize ($\pm2.01$)\\
& & & PolyNet$_{\text{in}}$ & 0.034 &\textbf{51.70 }\scriptsize ($\pm0.70$) &\textbf{33.54 }\scriptsize ($\pm0.74$) \\ 
\cdashline{4-7}
& & & FoMo-Net$_{1}$ & 66.39 & 42.87 \scriptsize($\pm1.69$) & 25.15 \scriptsize($\pm0.45$)  \\
& & & FoMo-Net$_{m}$ & 73.57 & 38.07 \scriptsize($\pm0.65$) & 19.44 \scriptsize($\pm0.45$)\\ \hline
\multirow{6}{*}{\multirowcell{2}{\textbf{TalloS} \\ \textbf{(ours)}}}
&  \multirowcell{6}{Multispec. \\ SAR \\ DEM}
& \multirow{6}{*}{\textbf{1160}}
& ResNet50 &26.34  & \textbf{27.31} \scriptsize($\pm1.02$) & \textbf{1.35} \scriptsize ($\pm0.09$) \\
& & & ViT &89.20  & 15.23 \scriptsize ($\pm3.97$) &  0.37\scriptsize ($\pm0.12$)\\
& & & ConvNext & 88.98 & 9.08 \scriptsize ($\pm2.58$)&0.43 \scriptsize($\pm0.12$)\\
& & & PolyNet$_{\text{in}}$ & 2.79 & 4.57 \scriptsize ($\pm0.88$) & 0.13 \scriptsize ($\pm0.03$)\\ 
\cdashline{4-7}
& & & FoMo-Net$_{1}$  &67.27 &  \underline{18.50} \scriptsize ($\pm 2.33$) & \underline{0.52} \scriptsize ($\pm0.15$)\\
& & & FoMo-Net$_{m}$ & 74.45 & 14.25 \scriptsize ($\pm2.81$) & 0.27 \scriptsize($\pm0.09$) \\ \hline
\multirow{6}{*}{\multirowcell{2}{\textbf{RapidAI4EO} \\ \cite{rapidai4eo}}}
&  \multirowcell{6}{\multirowcell{2}{Optical \\ Multispec.}}
& \multirow{6}{*}{\textbf{24}}
& ResNet50 &23.60 & \underline{62.84} \scriptsize ($\pm0.17$) & \underline{46.33} \scriptsize ($\pm0.81$) \\
& & & ViT & 88.37 & 51.04 \scriptsize ($\pm2.13$) & 25.59 \scriptsize ($\pm3.33$)\\
& & & ConvNext & 87.61 & \textbf{64.07} \scriptsize ($\pm0.23)$& \textbf{47.20} \scriptsize ($\pm5.24$) \\
& & & PolyNet$_{\text{in}}$ & 0.053 &59.21 \scriptsize ($\pm0.29$) &41.11 \scriptsize ($\pm0.40$) \\ 
\cdashline{4-7}
& & & FoMo-Net$_{1}$ & 66.40 & 54.69 \scriptsize ($\pm0.06$) & 33.47 \scriptsize ($\pm0.56$) \\
& & & FoMo-Net$_{m}$ & 73.58& 49.12  \scriptsize($\pm5.59$) & 24.63 \scriptsize ($\pm8.46$) \\ \hline 
\multirow{6}{*}{\multirowcell{2}{\textbf{TreeSatAI} \\ \cite{treesatai}}} 
&  \multirowcell{6}{Multispec.}
& \multirow{6}{*}{\textbf{15}}
& ResNet50 &23.56 & 53.01 \scriptsize ($\pm0.86$) & 30.19 \scriptsize ($\pm1.7$)\\
& & & ViT & 87.57 & \underline{55.48} \scriptsize ($\pm1.17$) & \underline{36.63} \scriptsize ($\pm1.99$) \\
& & & ConvNext & 87.60 & \textbf{64.15} \scriptsize ($\pm0.29$) & \textbf{ 43.99} \scriptsize ($\pm1.53$) \\
& & & PolyNet$_{\text{out}}$ & 0.03 & 42.41 \scriptsize ($\pm0.11$)& 21.41 \scriptsize ($\pm0.28$)  \\ 
\cdashline{4-7}
& & & FoMo-Net$_{1}$ &66.39 & 45.02 \scriptsize ($\pm 0.51$) & 18.93 \scriptsize ($\pm0.24$) \\
& & & FoMo-Net$_{m}$ &73.57 & 27.96 \scriptsize ($\pm1.02$) & 12.08 \scriptsize ($\pm0.41$)\\
\bottomrule
\end{tabular}
\end{center}
\caption{\textbf{Performance on classification tasks. }
 \#C denotes the number of classes and \#P the number of trainable parameters (in millions). We report the F1-score at the micro and macro levels. We \textbf{emphasize} the best results and \underline{underline} the second best.}
\label{table:perf_classif}

\end{table*}

\section{Experiments}

\label{sec:fomo_bench_eval}
We train a set of standard computer vision models tailored to each specific task. Each baseline is trained in a supervised fashion on the data for that single task due to the heterogeneity of the tasks in both input and output. 
%
%
In the case of multi-label classification and object detection, we also evaluate a simple approach using supervised learning on multiple datasets \ie PolyNet.
PolyNet denotes a ResNet-50 model pretrained on several datasets simultaneously. Since PolyNet has been trained in a supervised fashion on FoMo-Bench tasks, we evaluate it via linear probing either in-domain (denoted PolyNet$_{\text{in}}$), \ie on a dataset included in pre-training, or out-of-domain (denoted PolyNet$_{\text{out}}$), \ie on a dataset not included in pre-training. 
For object detection and semantic segmentation we train a task-specific head to learn the downstream task for each dataset independently. 
For all experiments, we present the average performance along with the standard deviation over three independent runs using different random seeds.
Detailed training settings and qualitative results are provided in Appendices C and D respectively.

\textbf{Semantic segmentation of grid-based data:} 
FoMo-Bench includes seven datasets suited to semantic segmentation. 
 Our baseline models are UNet \cite{ronneberger2015u}, DeepLabv3 \cite{chen2017rethinking} and UPerNet \cite{xiao2018unified}. 
 
\begin{table*}[!t]
\begin{center}\fontsize{8}{11}\selectfont
%
\begin{tabular}{c c c l c c c }
\toprule
Dataset & Modalities & \#C & Method & \#P (Mil.) & F1 micro & mIoU \\
\midrule
\multirow{5}{*}{\multirowcell{2}{\textbf{Waititu} \\ \cite{kattenborn2020convolutional}}}
& \multirow{5}{*}{Aerial RGB}
& \multirow{5}{*}{\textbf{3}}
& UNet &  32.52 & \underline{77.96} \scriptsize ($\pm3.36$) &  \underline{64.0} \scriptsize ($\pm4.55$)\\
& & & DeepLabv3 & 26.67   & \textbf{80.81} \scriptsize ($\pm2.81$) &\textbf{  67.90} \scriptsize ($\pm3.96$)\\
& & & UPerNet &  121,30  & 75.36 \scriptsize ($\pm3.04$) & 60.56 \scriptsize ($\pm3.94$)\\
\cdashline{4-7}
& & & FoMo-Net$_{1}$ &79.31 & 74.26 \scriptsize ($\pm2.00$) & 59.10 \scriptsize ($\pm 2.51$) \\
& & & FoMo-Net$_{m}$ & 86.49 & 71.52 \scriptsize ($\pm0.0$) & 55.66 \scriptsize ($\pm0.0$) \\
\midrule
\multirow{5}{*}{\multirowcell{2}{\textbf{FLAIR \#1} \\ \cite{ign2022flair1, garioud2023flair}}} 
& \multirow{5}{*}{Aerial multisp.}
& \multirow{5}{*}{\textbf{19}}
& UNet &   32.52 & \underline{72.80} \scriptsize ($\pm0.17$) &  \underline{57.24} \scriptsize ($\pm0.22$)  \\
& & & DeepLabv3 & 26.68   & \textbf{74.01 }\scriptsize ($\pm0.3$)&  \textbf{58.75} \scriptsize ($\pm0.39$) \\
& & & UPerNet &   121.30 & 63.89 \scriptsize ($\pm1.03$) & 46.94 \scriptsize ($\pm1.12$) \\
\cdashline{4-7}
& & & FoMo-Net$_{1}$ &86.39 & 66.16 \scriptsize ($\pm0.05$) & 49.43 \scriptsize ($\pm0.06$) \\
& & & FoMo-Net$_{m}$ &93.57 & 60.22 \scriptsize ($\pm0.39$) & 43.08 \scriptsize ($\pm0.39$)  \\ \hline
\multirow{5}{*}{\multirowcell{2}{\textbf{FLAIR \#2} \\ \cite{ign2023flair2, garioud2023flair}}} 
& \multirow{5}{*}{Aerial multisp.}
& \multirow{5}{*}{\textbf{19}}
& UNet &  32.56  & 72.99 \scriptsize ($\pm1.80$) & 57.50 \scriptsize ($\pm2.22$) \\
& & & DeepLabv3 &  26.7  & \textbf{83.97} \scriptsize ($\pm0.27$) &  \textbf{72.37} \scriptsize ($\pm0.40$)\\
& & & UPerNet & 121.33 &  \underline{76.93} \scriptsize ($\pm 0.28$) & \underline{62.52} \scriptsize ($\pm0.37$) \\
\cdashline{4-7}
& & & FoMo-Net$_{1}$ &121.78 & 64.69 \scriptsize($\pm0.07$) & 47.81 \scriptsize ($\pm0.08$) \\
& & & FoMo-Net$_{m}$ &128.96 & 57.95 \scriptsize($\pm0.17$) & 40.80 \scriptsize ($\pm0.16$)\\ \hline
\multirow{5}{*}{\multirowcell{2}{\textbf{Spekboom} \\ \cite{galuszynski2022automated}}} 
& \multirow{5}{*}{Aerial RGB}
& \multirow{5}{*}{\textbf{2}}
& UNet &  32.52  & \underline{97.70} \scriptsize ($\pm0.02$) &  \underline{95.50} \scriptsize ($\pm0.03$)\\
& & & DeepLabv3 & 26.67   &  97.57 \scriptsize ($\pm0.17$)& 95.27 \scriptsize ($\pm0.34$)\\
& & & UPerNet &  121.30  & \textbf{97.81 }\scriptsize ($\pm0.01$) & \textbf{95.72 }\scriptsize ($\pm0.03$)\\
\cdashline{4-7}
& & & FoMo-Net$_{1}$ & 79.31 & 95.59 \scriptsize ($\pm0.06$) & 91.56 \scriptsize ($\pm0.11$) \\
& & & FoMo-Net$_{m}$ & 86.49 & 91.83 \scriptsize ($\pm0.05$) & 70.97 \scriptsize ($\pm1.15$)\\ \hline
\multirow{5}{*}{\multirowcell{2}{\textbf{FiveBillionPixels} \\ \cite{FBP2023}}} 
& \multirow{5}{*}{Optical}
& \multirow{5}{*}{\textbf{22}}
& UNet &  32,52  & \textbf{74.40} \scriptsize ($\pm0.29$) & \textbf{59.24} \scriptsize ($\pm0.36$) \\
& & & DeepLabv3 & 26.68   & \underline{73.75} \scriptsize ($\pm0.32$) & \underline{58.42} \scriptsize ($\pm0.40$)  \\
& & & UPerNet &   121.31 & 67.43 \scriptsize ($\pm4.19$) & 51.02 \scriptsize ($\pm4.88$)\\
\cdashline{4-7}
& & & FoMo-Net$_{1}$ & 82.85 & 72.75 \scriptsize ($\pm0.19$) & 57.18 \scriptsize ($\pm0.24$)  \\
& & & FoMo-Net$_{m}$ & 90.03 & 67.39 \scriptsize ($\pm0.30$) & 50.82 \scriptsize ($\pm0.34$) \\ \hline
\multirow{5}{*}{\multirowcell{2}{\textbf{ForestNet} \\ \cite{irvin2020forestnet}}} 
&  \multirowcell{5}{Multispec.}
& \multirow{5}{*}{\textbf{5}}
& UNet &  32.52  & \underline{95.26} \scriptsize ($\pm0.02$) & \underline{90.96} \scriptsize ($\pm0.04$)\\
& & & DeepLabv3 & 26.68    & \textbf{95.27} \scriptsize ($\pm0.03$) & \textbf{90.97} \scriptsize ($\pm0.05$)\\
& & & UPerNet &  121.31  & 95.23 \scriptsize ($\pm0.05$) & 90.90 \scriptsize ($\pm0.10$)\\
\cdashline{4-7}
& & & FoMo-Net$_{1}$ & 79.31& 95.22 \scriptsize ($\pm0.02$) & 90.89 \scriptsize ($\pm0.04$) \\
& & & FoMo-Net$_{m}$ & 86.49& 95.24 \scriptsize ($\pm0.0$) & 90.92 \scriptsize ($\pm0.0$)  \\ \hline
\multirow{5}{*}{\multirowcell{2}{\textbf{Woody} \\ \cite{kattenborn2019uav}}} 
& \multirow{5}{*}{Aerial RGB}
& \multirow{5}{*}{\textbf{4}}
& UNet &  32.52   & \underline{93.37} \scriptsize ($\pm0.23$) & \underline{87.57} \scriptsize ($\pm0.42$)\\
& & & DeepLabv3 &  26.67  & 93.30 \scriptsize ($\pm0.40$) & 87.44 \scriptsize ($\pm0.71$)  \\
& & & UPerNet &   121.30  & \textbf{93.87} \scriptsize ($\pm0.17$)& \textbf{88.46} \scriptsize ($\pm0.31$)\\
\cdashline{4-7}
& & & FoMo-Net$_{1}$ &79.31 & 89.7 \scriptsize ($\pm0.04$) & 81.32 \scriptsize($\pm0.06$)  \\
& & & FoMo-Net$_{m}$ &86.49 & 84.33 \scriptsize ($\pm0.84$) & 72.93 \scriptsize ($\pm1.27$) \\ \hline
\bottomrule
\end{tabular}
\end{center}
\caption{\textbf{Semantic segmentation tasks for grid-based data.} 
\#C denotes the number of classes and \#P the number of classes and number of trainable parameters (in millions). We report the F1-score and the mean intersection over union (mIoU) at the micro levels. We \textbf{emphasize} the best results and \underline{underline} the second best.}
\label{table:perf_seg_img}
 \end{table*}
For UNet and DeepLabv3 architectures, we use a ResNet50 \cite{he2016deep} backbone pretrained on ImageNet. The UPerNet is based on a Swin-Base backbone from \cite{mmseg2020}. 
We assess FoMo-Net by extracting the encoded tokens, upsampling and feeding them to a small decoder similar to FloodViT~\cite{bountos2024kurosiwo33billion}. 
We report the mean intersection over union (mIoU) and the F1-Score at micro level for all datasets. 
Results are presented in Tab.~\ref{table:perf_seg_img}. 

\textbf{Classification:}
Due to the large GSD of sensors in RS, classification tasks are usually multi-label.
The multi-label classification setting is thus used for BigEarthNet-MM, Sen12MS, RapidAI4EO, TreeSat AI and TalloS for either LULC or tree species recognition.
Our baselines are ResNet50, ViT and ConvNext \cite{liu2022convnet} models. We initialize all models with weights pretrained on ImageNet. 
We pretrain PolyNet on 
RapidAI4EO, TalloS, Sen12MS and BigEarthNet, with their common bands. 
In-domain evaluation is performed on these datasets, while TreeSatAI is used for out-of-domain evaluation.
We report micro and macro F1 scores for all datasets in Tab.~\ref{table:perf_classif}.

\textbf{Object detection}
The object detection task is useful in forest monitoring to count and precisely locate trees, focusing in some cases on particular species.
We use the NeonTree and ReforesTree datasets to test binary and multi-class object detection respectively, testing the Faster R-CNN \cite{faster_rcnn}, RetinaNet \cite{retinanet} and YOLOS \cite{yolos} models (see Tab.~\ref{table:perf_detection}). We measure performance by the mIoU, the mean average precision with an $\text{IoU}=0.5$ threshold ($\text{mAP}_{50}$) and the mean average precision with a variable threshold of $\text{IoU} \in [0.5, 0.95]$ ($\text{mAP}_{50:95}$).

%
The models include ResNet-50 and DeiT-S \cite{pmlr-v139-touvron21a} backbones pre-trained on the COCO dataset. 
We also include a Faster R-CNN with a DINOv2 backbone \cite{oquab2023dinov2} fully finetuned.
FoMo-Net is augmented with a randomly initialized Faster R-CNN head for finetuning.
We pretrain PolyNet simultaneously on NeonTree and ReforesTree. In-domain evaluation (PolyNet$_{\text{in}}$) is then conducted on each dataset separately. For the out-of-domain evaluation (PolyNet$_{\text{out}}$), we pretrain the backbone on one dataset while evaluating on the other.

\textbf{Semantic segmentation of point clouds:} 
Processing point cloud representations from LiDAR sensors aims at better understanding the geometry of trees at large scale. 
FoMo-Bench includes NeonTree and FORinstance datasets for binary and multi-class point cloud segmentation respectively.
For these datasets, we use the baselines PointNet \cite{qi2017pointnet}, PointNet++ \cite{qi2017pointnet2} and Point Transformer \cite{Zhao_2021_ICCV}. 
These experiments are presented in Appendix E.

\section{Discussion}
\label{sec:discussion}
\begin{table*}

\begin{center}\fontsize{8}{11}\selectfont
\begin{tabular}{c c c l l c c c c }
\toprule
Dataset & Modality & \#C & Method & Backbone & \#P (Mil.) & mIoU & $\text{mAP}_{50}$ & $\text{mAP}_{50:95}$ \\
\midrule
\multirow{8}{*}{\multirowcell{2}{\textbf{NeonTree} \\ \cite{weinstein2020benchmark}}}
& \multirow{8}{*}{Airborne}
& \multirow{8}{*}{\textbf{1}}
& F. R-CNN & ResNet-50 & 41.08  & \textbf{61.49}	\scriptsize ($\pm0.087$) & \textbf{38.37} \scriptsize ($\pm1.68$) & \textbf{13.72} \scriptsize ($\pm0.68$)  \\
& & & RetinaNet & ResNet-50 & 36.13  & \underline{56.27} \scriptsize ($\pm1.45$) & 6.42 \scriptsize ($\pm4.37$)& 2.10 \scriptsize ($\pm1.36$)   \\
& & & YOLOS & DeiT-S &  30.65  & 53.36 \scriptsize ($\pm0.18$) & 28.34 \scriptsize ($\pm0.48$) & \underline{12.64} \scriptsize ($\pm0.31$)  \\
& & & F. R-CNN & PolyNet$_{\text{in}}$ &  14.50  & 58.42 \scriptsize ($\pm0.04$) & \underline{35.93} \scriptsize ($\pm0.10$) & 11.95 \scriptsize ($\pm0.03$) \\
& & &  F. R-CNN & PolyNet$_{\text{out}}$ &  14.50  & 56.05 \scriptsize ($\pm0.10$) & 27.77 \scriptsize ($\pm0.21$) & 8.70\scriptsize ($\pm0.07$) \\
& & &  F. R-CNN & DINOv2 &  94.36  & 44.65 \scriptsize ($\pm0.65$) & 1.08 \scriptsize ($\pm0.73$) & 0.22 \scriptsize ($\pm0.17$) \\
\cdashline{4-9}
 & & & F. R-CNN & FoMo-Net$_{1}$  & 81.37  & 51.95 \scriptsize ($\pm 0.46$)  & 5.67 \scriptsize ($\pm0.82$) & 1.52 \scriptsize ($\pm0.25$) \\
& & & F. R-CNN & FoMo-Net$_{m}$ & 88.55  &45.12 \scriptsize ($\pm 0.42$)  & 1.1 \scriptsize ($\pm0.56$)  &0.24 \scriptsize ($\pm0.13$) \\
\midrule
\multirow{8}{*}{\multirowcell{2}{\textbf{ReforesTree} \\ \cite{reiersen2022reforestree}}} 
& \multirow{8}{*}{Aerial}
& \multirow{8}{*}{\textbf{6}}

& F. R-CNN & ResNet-50 & 41.10 & \underline{64.67} \scriptsize ($\pm0.78$) & 6.83 \scriptsize ($\pm2.33)$ & \underline{2.91} \scriptsize ($\pm1.11$)  \\
& & & RetinaNet & ResNet-50 & 36.21 & \textbf{65.49} \scriptsize ($\pm0.45$) & 4.69 \scriptsize ($\pm1.02$) & 2.13 \scriptsize ($\pm0.53$)  \\
& & & YOLOS & DeiT-S &  30.65  & 62.85 \scriptsize ($\pm0.60$) &  2.30  \scriptsize ($\pm0.49$) & 1.33 \scriptsize ($\pm0.23$) \\
& & & F. R-CNN & PolyNet$_{\text{in}}$ &  14.52  & 62.87 \scriptsize ($\pm0.25$) & \textbf{8.14} \scriptsize ($\pm0.11$) &\textbf{ 3.30} \scriptsize ($\pm0.03$) \\
& & & F. R-CNN & PolyNet$_{\text{out}}$ &  14.52 & 59.47 \scriptsize ($\pm0.48$) & \underline{7.19} \scriptsize ($\pm0.22$) & 2.73 \scriptsize ($\pm0.09$) \\
& & & F. R-CNN & DINOv2 &  94.38  & 46.90 \scriptsize ($\pm0.36$)  & 0.08 \scriptsize ($\pm0.02$) & 0.02 \scriptsize ($\pm0.01$)  \\
\cdashline{4-9}
& & & F. R-CNN & FoMo-Net$_{1}$  & 81.39  & 51.63 \scriptsize ($\pm 0.25$)  & 1.32 \scriptsize ($\pm0.50$)  & 0.46  \scriptsize ($\pm0.16$) \\
& & & F. R-CNN & FoMo-Net$_{m}$ & 88.57  & 46.38 \scriptsize ($\pm 0.24$)  & 0.20 \scriptsize ($\pm0.08$)  & 0.04 \scriptsize ($\pm0.02$) \\
\bottomrule
\end{tabular}
\end{center}
\caption{\textbf{Object detection evaluation.} 
\#C denotes the number of classes and \#P number of trainable parameters  (in millions). 
We report the mIoU and the mean average precision (mAP) for both $\text{IoU}=0.5$ and $\text{IoU} \in [0.5, 0.95]$ thresholds. We \textbf{emphasize} the best results and \underline{underline}
the second best.} 
\label{table:perf_detection}
\end{table*}
 \textbf{FoMo-Bench:} Interestingly, traditional CNNs consistently rank among the top-performing models in both classification and segmentation tasks (Tab.~\ref{table:perf_seg_img},~\ref{table:perf_classif}). ResNet50 exceptional performance on TalloS is striking, with FoMo-Net$_{1}$ coming close. TalloS high number of classes along with their skewed distribution make it an extreme multilabel classification problem, reflecting the complexities of global-scale forest monitoring. Tree species are distributed unequally around the globe (see Fig.~\ref{fig:tallos_distr}), with many classes present only in specific locations. Incorporating tree species occurrence maps as biological prior knowledge could prove helpful. 
PolyNet's performance raises interesting research questions on the utilization of the semantic information of forest monitoring datasets for the creation of foundation models. With the exception of TalloS, where it is severely outperformed by FoMo-Net, PolyNet shows good in-domain predictive skill, including attaining the best performance for Sen12MS. 
FoMo-Net pretraining yielded promising results, even though it is still outperformed by specialized models. Notably, in some cases it surpasses the original ViT, particularly in TalloS and RapidAI4EO. 
As discussed in Sec.~\ref{sec:fomo-net}, digesting such diverse information simultaneously might benefit from a larger encoder. Despite the lack of built-in functionality in ViT 
for processing hierarchical features in dense prediction, FoMo-Net's performance in segmentation tasks (Table ~\ref{table:perf_seg_img}) is highly competitive to supervised specialized architectures \eg UPerNet. Although the supervised baselines in FoMo-Bench set a high bar for comparison, they are not directly comparable to models like FoMo-Net. While supervised-learning models can excel when trained on large datasets, they struggle when applied to different tasks and are fundamentally limited to the specific tasks they were trained on. FoMo-Net pushes the boundaries of this emerging field, establishing a basis for the development of future foundation models. Another key finding is the superiority of FoMo-Net$_1$ over FoMo-Net$_m$. Contrary to intuition, a single shared projection to the embedding space outperforms dedicated projections for each spectral band. This enhances FoMo-Net's flexibility as new modalities can rely on the shared embedding. We compare FoMo-Net with other foundation models in Appendix F. 

\textbf{Limitations and next steps:} While FoMo-Bench offers strong baselines across a broad range of forest monitoring tasks, assessing the true capability of flexible foundation models requires more than just performance metrics. Given the diversity in input modalities 
we need methods to evaluate how well these flexible models recognize the unique characteristics of each modality, their inter-modal relationships, and the features of the Earth's surface across multiple scales. 
Also, as FoMo-Net tokenizes spectral bands independently the amount of tokens per sample increases proportionally to the number of spectral bands, creating a computational bottleneck due to the memory requirements and the quadratic complexity of attention posing challenges for scalability and making optimization a priority. Given PolyNet's potential, combining FoMo-Net with supervised pre-training might result in a flexible pre-training method usable under variable computational constraints and with variable amounts of labeled data. 
Finally, extending FoMo-Net to non-grid data, \eg point clouds,
could prove valuable. 

\section{Conclusion}
This work presents FoMo-Bench, a global benchmark, providing a robust evaluation framework for future forest monitoring flexible foundation models.  To further expand the tasks and spatial coverage of FoMo-Bench, we introduce TalloS, a global, multi-temporal and multi-modal dataset for tree species classification. Building on these contributions, we craft a pre-training framework to construct flexible foundation models \ie FoMo-Net, and provide an initial assessment on FoMo-Bench identifying current limitations and potential opportunities for this promising field. 
\appendix
\vspace{.2em}



\section*{Acknowledgments}
This work has received funding from the IVADO program on “AI, Biodiversity and Climate Change”, the Canada CIFAR AI Chairs program and the European Union's Horizon Europe research and innovation program under grant agreement No. 101130544 (ThinkingEarth).
\bibliography{aaai25}

\end{document}